\documentclass[conference]{IEEEtran}
\ifCLASSINFOpdf
\else
\fi
\usepackage{adjustbox}

\usepackage{orcidlink}
\usepackage{amsmath}
\usepackage{amssymb}
\usepackage{algorithm}
\usepackage{algpseudocode}
\usepackage{booktabs} 
\usepackage{pifont}   
\usepackage{amssymb}  
\usepackage{adjustbox}

\newtheorem{remark}{Remark}

\begin{document}
%
\title{GEM-RAG: Graphical Eigen Memories For Retrieval Augmented Generation}


\author{\IEEEauthorblockN{Brendan Hogan Rappazzo, Yingheng Wang, 
Aaron Ferber, Carla Gomes}
\IEEEauthorblockA{Department of Computer Science\\
Cornell University\\
Ithaca, New York 14850\\
Email: \{bhr54, yw2349, amf272, cpg5\}@cornell.edu}}

\maketitle

\begin{abstract}
The ability to form, retrieve, and reason about memories in response to stimuli serves as the cornerstone for general intelligence - shaping entities capable of learning, adaptation, and intuitive insight. Large Language Models (LLMs) have proven their ability, given the proper memories or context, to reason and respond meaningfully to stimuli. However, they are still unable to optimally encode, store, and retrieve memories - the ability to do this would unlock their full ability to operate as AI agents, and to specialize to niche domains. To remedy this, one promising area of research is Retrieval Augmented Generation (RAG), which aims to augment LLMs by providing them with rich in-context examples and information. In question-answering (QA) applications, RAG methods embed the text of interest in chunks, and retrieve the most relevant chunks for a prompt using text embeddings. Motivated by human memory encoding and retrieval, we aim to improve over standard RAG methods by generating and encoding higher-level information and tagging the chunks by their utility to answer questions. We introduce Graphical Eigen Memories For Retrieval Augmented Generation (GEM-RAG). GEM-RAG works by tagging each chunk of text in a given text corpus with LLM generated ``utility'' questions, connecting chunks in a graph based on the similarity of both their text and utility questions, and then using the eigendecomposition of the memory graph to build higher level summary nodes that capture the main themes of the text. As a result, GEM-RAG not only provides a more principled method for RAG tasks, but also synthesizes graphical eigen memory (GEM) which can be useful for both exploring text and understanding which components are relevant to a given question. We evaluate GEM-RAG, using both UnifiedQA and GPT-3.5 Turbo as the LLMs, with SBERT, and OpenAI's text encoders on two standard QA tasks, showing that GEM-RAG outperforms other state-of-the-art RAG methods on these tasks. We also discuss the implications of having a robust RAG system and future directions. 
\end{abstract}


%
\IEEEpeerreviewmaketitle

\section{Introduction}
\label{sec:intro}The ability to create intelligent machines has long occupied the fascination of humankind, from the automata of the medieval error, formalization of logic in the 17th century and evolving with the emergence of computing theory and artificial intelligence concepts in the 19th century. Now in the modern era, the possibility of generating Artificial General Intelligence (AGI) \cite{sparks}  seems as close as ever, in particular just within the past three years with the advent of massive scale machine learning systems, especially Large Language Models (LLMs), and Large Multimodal Models (LMMs). 

LLMs have emerged as remarkably powerful general knowledge stores, with the ability to perform impressively on a large variety of tasks \cite{knowledge_1_petroni_language_2019,knowledge_2DBLP:journals/corr/abs-1911-12543,bubeck2023sparks}. Further, given their large context lengths, its has been shown that they have a powerful ability to perform in-context learning, whether it be for chat-like applications where it can reference parts of the conversation dynamically, adapting to classification tasks \cite{in_context_raventós2023pretraining}, or reasoning tasks through chain-of-thought prompting \cite{tree_yao2023tree}. LLMs appear to have solved one part of the general intelligence equation, given the proper context, much like a human mind given the proper working memory stream, they can reason and respond to questions reasonably. But without a way to encode, store and retrieve information that extends outside of their context, LLMs are missing an extremely important part of general intelligence, they are unable to form long-term, and ongoing memories as AI agents, and are unable to adapt to new niche domains. 

If LLMs were able to perfectly encode, store and retrieve memories it would open the possibility for AI agents to remember decades of conversations, research publications, literary works, and more, building hierarchies of memories and knowledge about the world. It would allow advanced QA models to read and encode huge amounts of niche text documents, and provide robust conversations or QA citing content in those documents. Expanding this idea for LMMs, this would enable the ability to retrieve text, memories, audio, imagery, video, and more, paving the way for more potent models and use cases.




\begin{figure*}[t]
    \centering
    \includegraphics[width=1.0\textwidth]{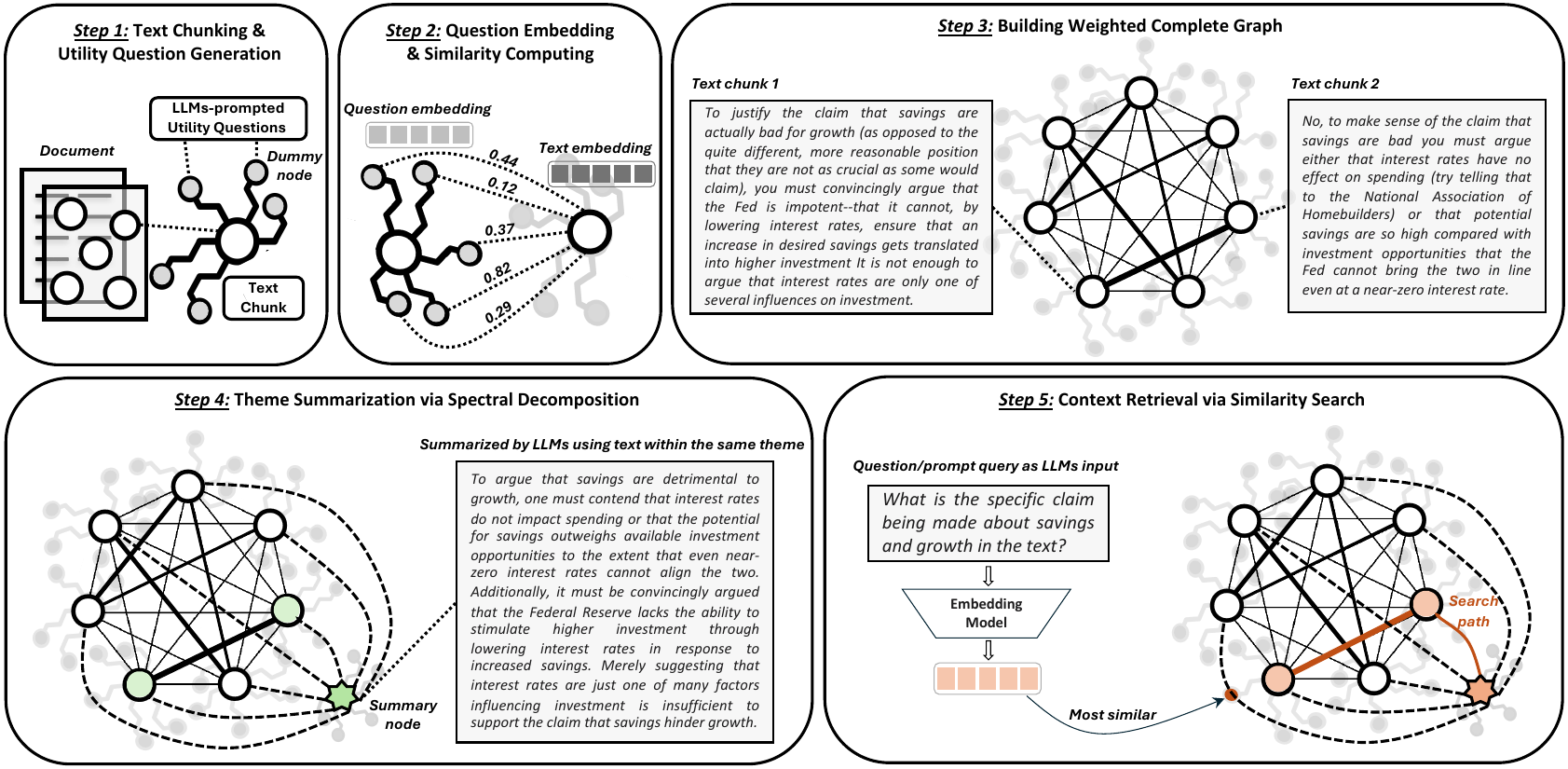}
    \caption{An overview of the graph construction and retrieval process for GEM-RAG. Given a corpus of text, the text is first grouped into chunks of text. GEM-RAG then generates utility questions for each chunk of text using an LLM, where a utility question asks something that could be answered given the text chunk as context. Next GEM-RAG uses these utility questions and respective embeddings to build a weighted graph. Summary nodes are then generated using the graph's spectral decomposition, using the eigenvectors to represent different orthogonal modes or ``eigenthemes'' of the text. For retrieval, GEM-RAG embeds the question or prompt, and searches the graph for the optimal nodes or context to return.}
    \label{fig:schematic}
\end{figure*}


Motivated by this problem, retrieval augmented generation (RAG) has been an increasingly active area of research, which aims to adapt static LLMs to new and niche domain applications \cite{rag_1,rag_2_DBLP:journals/corr/abs-2005-11401,rag_3_DBLP:journals/corr/abs-1804-09541,rag_4_ram2023incontext,rag_5_izacard2022atlas}. While RAG methods can work for any type of text-based information, whether it  be memories \cite{park2023generative} or images in the case of vision LMMs \cite{ragdriver}, one standard setting is that concerning large corpora of text from stories, articles, books, etc. In these settings, RAG generally works by first splitting the text into chunks, and obtaining an embedding for each chunk. Given a new prompt, the RAG system will embed it, return the top $k$ nearest chunks in embedding space, and then prompt the LLM to answer the prompt given the context chunks \cite{tracing_akyürek2022tracing}.

Current RAG systems empirically work reasonably well, but are far from optimal and have several issues, of which we will discuss two. Firstly, while it is a good approximation, purely finding the similarity between the prompt and each chunk in the embedding space may not be the best method to retrieve the most relevant chunks to answer a question. The texts may be similar in style or phrasing, giving a superficially high similarity score, while the context or purpose of the text unrelated to the question. Secondly, often answering complex questions requires a synthesis of information across many chunks of text, and understanding how different chunks are interrelated and connect to each other conceptually is important for retrieval.  

Aiming to solve these problems, we took inspiration from human cognition, specifically how humans encode and retrieve information based on the information's relevance and utility \cite{cognitive_psychology}. For instance, information tends to be better encoded if the person is tested on the material \cite{testing_encoding}, suggesting the importance of understanding the utility of the encoded information. Further, the more often memories are retrieved together, the more humans will synthesize this data into higher level summaries \cite{retrieval}. 

Motivated by human cognition, we propose Graphical Eigen Memories For Retrieval Augmented Generation (GEM-RAG). Given a corpus of text, our method splits the text into chunks and then generates several relevant ``utility'' questions using an LLM. The embeddings of these utility questions help build a complete weighted graph, where the weight between two nodes is the similarity of the utility questions. With this memory graph, and motivated by the observation that humans tend to synthesize information that is often retrieved together, we perform a random-walk analysis on the graph, by using the spectral decomposition of the normalized graph Laplacian. The intuition is that the eigenvectors of this decomposition provides an understanding of the key orthogonal themes present in the passage, by providing the different modes based on memory node similarity. We use the top components of important eigenvectors to then produce ``eigenthemes'' or summary nodes, which capture the higher-level structure of the text. This synthesized graph constitutes the graphical eigen memory, or GEM. At retrieval time, we find the most similar utility question to the prompt/question, and then perform best first search on the GEM, to retrieve a sub-graph containing the context chunks ultimately passed to the LLM. A schematic of our method can be seen in Figure \ref{fig:schematic}.

We believe developing a robust RAG system will enable LLMs to become rich AI-agents, adapting smartly to niche scientific domains, and unlock other important applications dependent on actively using memory. To best constrain and study the effectiveness of our method, we quantitatively evaluate the accuracy of our model on two QA datasets, QuALITY \cite{pang2022quality} and Qasper \cite{qasper_dasigi-etal-2021-dataset}. We compare our method against a baseline of the standard RAG procedure, as well as against a recent state-of-the-art method, RAPTOR \cite{raptor}. For each method we explore using the SBERT \cite{sbert} and OpenAI's text-embedding-ada-002 embedding models, as well as the UnifiedQA \cite{unifiedqa-test} and GPT-3 \cite{gpt3} as LLMs. We show that in most cases our method gives better performance, and we provide ablation experiments to give an understanding of the effect of the number of eigen summary nodes and utility questions. Lastly, while the primary use-case of the synthesized graphical eigen memory (GEM) is to perform RAG for an LLM, we also explore how the GEM is a standalone object agnostic to the LLM. Ultimately, the GEM can be used to explore and visualize data, which we showcase in a web demo for an example graph. 

Our contributions are as follows, \textbf{1)} We introduce a novel RAG system inspired by human cognition that encodes, store and retrieves information by its utility. \textbf{2)} We further formulate generating summary nodes as a random walk problem, and use eignedecomposition to generate summary nodes. \textbf{3)} We demonstrate the effectiveness of our RAG method on two QA datasets, using multiple different embedding and language models. We ran several ablation experiments to better understand the effectiveness of our method. \textbf{4)} We release an interactive web demo of an example GEM, to showcase how the graph works, and to emphasize its use as a standalone object.

\section{Related Work}
\textbf{Large Language Models (LLMs)}
LLMs have proven to be extremely powerful general knowledge stores \cite{gpt_3, unifiedqa-test, openai2023gpt4}. It has also been shown in some cases, that fine-tuning can produce data specific models \cite{dpr}. The advent of better hardware and algorithms has allowed for them to have larger context lengths, which can handle more information to be learned or retrieved in-context. However, it has been shown that longer contexts have diminishing returns, can lead to a loss of information \cite{liu2023lost,sun-etal-2021-long}, still necessitating the use of retrieving relevant contextual information.

It has also been shown, that with large context length models can perform very well on short story QA tasks \cite{zeroscrolls}. However, it should be noted that these methods retrieve the entire story as context, with thousands of tokens, where as we study only returning hundreds, as is the case in RAG methods. The motivation for this is discussed in the experimental section. 


\textbf{Retrieval Augmentation Methods}
The area of retrieval augmented generation has seen recent interest. In the work \cite{rag_1} initially introduced the idea of augmenting the context of an LLMs with the retrieved information. Further work shown  by \cite{rag_2_DBLP:journals/corr/abs-2005-11401,rag_3_DBLP:journals/corr/abs-1804-09541,rag_4_ram2023incontext,tracing_akyürek2022tracing} expanded on the idea, conceptualizing how LLMs could use retrieved context to trace where its response was coming from. In the work \cite{izacard2022atlas} proposed learning the retriever and LLM model jointly where as another work proposed using a tree-decoding algorithm for multi-answer retrieval \cite{min-etal-2021-joint}. Many works use different hierarchies of data summary \cite{arivazhagan-etal-2023-hybrid, liu-etal-2021-dense-hierarchical}. The work in \cite{wu2021recursively} also showed how hierarchies can be built using recursive summarization. In the work \cite{newman2023question} showed that RAG methods, while improving over baselines, often fail to provide enough key context to properly answer the question. It has also been showed how to use LLMs to generate summaries of chunks of text for improved accuracy. And the benefits of doing RAG with a custom encoder module \cite{gao2023enabling}, \cite{karpukhin-etal-2020-dense} Recently, Recursive Abstrative Processing
For Tree-Organized Retrieval \cite{raptor}, studied the effects of building hierarchical trees of chunk summaries based on aggregating textual chunks by similarity. However, this approach fails to consider the biases of textual similarity and limits the node synthesis to a tree structure. 

\section{Methods}
\subsection{Connection to Human Cognition}
GEM draws its design and motivation from the intricacies of human cognition, namely, the processes by which humans encode, store and retrieve information. Specifically, we draw inspiration from how it is believed the human brain prioritizes information based on its utility, and that information that is most often retrieved together gets summarized together \cite{cognitive_psychology}.

Our first observation from psychology is the so called ``testing effect", which observes that if humans are tested on subject material, they are more likely to accurately remember it \cite{testing_encoding}. This may be because, by testing, the information can be associated to a specific utility in our cognition, and thus is ``tagged" with information that makes it easier to retrieve. To this end, we aim to better tag each chunk of information, by ``tagging" it with LLM generated utility questions, which help express what information a specific text chunk has, and why it might be useful. 

Our second observation in that in human cognition, the more often memories are retrieved together, the more likely the will continue to be retrieved together \cite{retrieval}. In the context of our utility questions method and their respective text embeddings, we can build a graph constituting all chunks of the text as nodes, and the strength of their connections as the similarity of their question embeddings. We can then, inspired by this observation of human cognition, try to build higher level summary nodes based off the modes of the graph, i.e. the nodes that are likely to be retrieved together given a prompt. Our intuition is that by performing an eigendecomposition, each eigenvector will capture different mode or themes in the text, that would often be retrieved together, and thus should be used to generate summary nodes.

\subsection{Graphical Eigen Memories}
Our method at a high level involves several steps to first construct the graph, by chunking the text, generating utility questions, building the initial graph, and finally using the eigenvectors from spectral decomposition of normalized graph Laplacian to build the summary nodes. Then, with the GEM produced, we show how it can be used to perform RAG. The specifics of our implementation are discussed in the experiments section. A schematic of our method can be seen in Figure \ref{fig:schematic}.

\subsubsection{Memory Graph Construction}
\textbf{Chunking}
Text chunking is standard practice in RAG where the corpus of text of interest $C$ is split apart into chunks, $\left\{\mathtt{CHUNK}_1\ldots \mathtt{CHUNK}_n\right\}$, where each chunk, $\mathtt{CHUNK}_i$ has some number $T$ tokens. In practice the text can be split by number of characters or number of tokens. For a corpus of text we first chunk the text into $n$ chunks, where each chunk is $T$ tokens long, and $n=\frac{N}{c}$, where $N$ is the total number of tokens in the text.

\textbf{Generating Utility Questions}
Given each text $\mathtt{CHUNK}_i$, we then prompt an LLM to generate some $m$ number of utility questions. This can be represented by a function $Q$, that takes a chunk of text, and an integer, $m$, and generates $m$ utility questions. Formally, given $\mathtt{CHUNK}_i$, we compute $Q(\mathtt{CHUNK}_i, m)$ which give a set of utility questions $ \{q_{i1}, q_{i2}, \ldots, q_{im}\}$.

\textbf{Text Embeddings}
In order to quantify the similarity of each utility question to each other and/or to a prompt, we need to embed the text into a high dimensional feature space, given by a text encoder. More specifically, given a text embedding function $E$,  for each node $i$, we compute the embedding $v$ as follows: $v_{i} = E(\mathtt{CHUNK}_i)$. 
Similarly, we also compute this embedding for each utility question by the same function. However, in order to best encode the information of the utility question we make the utility question's embedding the average of the questions embedding, and the base text's embedding, that is $v_{ij} = (E(q_{ij})+v_{i})/2$. 

\textbf{Building the Weighted Complete Graph}
Given the embedding of each chunk of text, as well as each chunk's corresponding utility questions and respective embeddings, we then generate the fully connected graph. For each node/chunk pair $i,j$, we consider the sum of the similarity metric between all of node $i$'s utility question embeddings, to that of node $j$'s base text embedding. More formally, let $G = (\mathcal{N},E)$ be the memory graph we are constructing. Let the nodes be given by: $\mathcal{N} = \{\mathtt{CHUNK}_1, \mathtt{CHUNK}_2, \ldots, \mathtt{CHUNK}_K\}$, where each node $t$ has utility questions $Q_t = \{q_{t1}, q_{t2}, \ldots, q_{tm}\}$, and base text embedding $v_{t}$. Then, $\text{For each } (t, v) \in \mathcal{N} \times \mathcal{N}, \text{ and for each } i \in \{1, 2, \ldots, m\},
\text{generate an edge between } t \text{ and } v$  with weight  $\sum^{m}_{i=0}\mathtt{SIM}(v_{ti}, v_{v})$. Any function could be used to compute the similarity, but in all cases we use standard cosine similarity, i.e., $\mathtt{SIM}(a, b) = \frac{a \cdot b}{\|a\| \|b\|}$.

\textbf{Building the Summary Nodes}
In order to build an encoding system that encodes this higher level information we formulate this as a random walk or spectral decomposition problem. Intuitively, in this context, each eigenvalue and its corresponding eigenvector reveal a distinct `theme' or conceptual dimension in the graph. By summarizing the top component nodes of each eigenvector components, using an LLM, we can understand the most significant relationships and conceptual clusters within the graph.

More specifically, let $ S = (s_{ij})_{n \times n}$ be the similarity matrix of the graph, where $s_{ij}$ is the sum weight between the $m$ $\mathtt{SIM}$ values of each utility questions of nodes $i$ and to node $j$. By attaching each $s_{ij}$ to a weighted edge $e_ij$, we can map the similarity matrix $S$ onto the memory graph $G$. Thus, we can better understand the relationship between different text pieces of the document by analyzing the properties and behaviors of $G$. 

Since different documents possess different connectivity and node centrality, the spectrums will also be at different scales. To better quantify how influential each node is without degree bias, we transform $S$ to a variant of normalized graph Laplacian $L$, which is $L = D^{-1/2} (S - I) D^{-1/2}$, where $D = \text{diag}(d_i)$ is the degree matrix and $I$ is the identity matrix. Then we conduct spectral decomposition by solving the following $ L\vec{x} = \lambda\vec{x}$.
With the resulting eigenvalues $\lambda_1, \lambda_2, \ldots, \lambda_n$ ordered in non-increasing order of their magnitude. The corresponding eigenvectors $\vec{x}_1, \vec{x}_2, \ldots, \vec{x}_n $  represent the principal themes. Then, for each eigenvector $\vec{x}_k$, select the top $e$ components, $x_{k1}, x_{k2}, \ldots, x_{ke}$, representing the most relevant nodes for the $k$-th theme. Then we prompt an LLM, given the text passages associated with $x_{k1}, x_{k2}, \ldots, x_{ke}$, to summarize the text, summarizing the high-level information. With the produced summary text, we introduce it as a new node in the graph. Lastly, as in previous steps, we produce utility questions, embeddings for the utility questions, and connect it to every other node in the graph in the manner previously described. 

\begin{algorithm}[!t]
    \caption{Retrieval from GEM-RAG}
    \textbf{Input}: \(p\) (prompt), \(\mathcal{B}\) (budget)\\
    \textbf{Output}: \(\mathcal{C}\) (set of context nodes)
    \begin{algorithmic}[1]
        \State \(v_p \gets E(p)\) \Comment{Embed the prompt \(p\) using the embedding function \(E\)}
        \State \(\mathcal{Q} \gets \text{Set of all utility question embeddings}\)
        \State \(\mathcal{C} \gets \{\emptyset\}\) \Comment{Initialize retrieval set \(\mathcal{C}\) to the empty set.}
        \While{\(|\mathcal{C}| < \mathcal{B}\)} 
            \Comment{Expand the set until the budget \(\mathcal{B}\) is reached}
            \State \(n = \underset{n_i \in \mathcal{Q} \setminus \mathcal{C}}{\mathrm{argmax}}\, \mathtt{SIM}(v_p, v_{ni})\)
            \State \(\mathcal{C} \gets \mathcal{C} \cup \{n\}\) \Comment{Add the most relevant node \(n\) to the set \(\mathcal{C}\)}
        \EndWhile
        \State \Return \(\mathcal{C}\) \Comment{Return the set \(\mathcal{C}\) as the context for the prompt \(p\)}
    \end{algorithmic}
    \label{alg:retrieval}
    
\end{algorithm}
\textbf{Analysis of Graph Spectrum}

From spectral graph theory \cite{chung1997spectral, li2007enhancing}, the eigenvalues of $L$ exhibit these properties: (1) $\sum \lambda_i = 0$, (2) each eigenvalue $\lambda_i$ falls within the range $[-1, 1]$, and (3) the largest eigenvalue $\lambda_1$ is 1. The similarity matrix $S-I$, with zero diagonals, has $(n^2 - n)$ non-zero elements, leading to an $O(n^2)$ complexity for the normalized Laplacian transformation. Subsequent spectral decomposition via the Lanczos algorithm \cite{Lanczos} incurs an $O(n^3)$ computational complexity.

Given that all text chunk nodes are clustered into different themes via spectral decomposition, we can observe some interesting properties on $S$ from such clustering behavior. 

\begin{table*}[t]
\centering
\begin{tabular}{lcccc}
\toprule
Embedding & LLM   & RAG    & Acc  & {HARD Acc} \\
\midrule
SBERT     & UnifiedQA & GEM-RAG   & \textbf{52.14}\% & \textbf{44.70}\% \\
SBERT     & UnifiedQA & RAPTOR & 51.71\% & 43.30\% \\
SBERT     & UnifiedQA & Embed  & 51.04\% & 44.06\% \\
\midrule
SBERT     & GPT3.5& GEM-RAG   & \textbf{61.84}\% & \textbf{51.60}\% \\
SBERT     & GPT3.5& RAPTOR & 60.13\% & 50.32\% \\
SBERT     & GPT3.5& Embed  & 58.61\% & 47.25\% \\
\midrule
OpenAI    & UnifiedQA & GEM-RAG   & 52.81\% & 44.83\% \\
OpenAI    & UnifiedQA & RAPTOR & \textbf{53.48}\% & \textbf{44.96}\% \\
OpenAI    & UnifiedQA & Embed  & 52.14\% & 42.53\% \\
\midrule
OpenAI    & GPT3.5& GEM-RAG   & \textbf{63.37}\% & \textbf{51.85}\%\\
OpenAI    & GPT3.5& RAPTOR & 60.32\% & 50.96\% \\
OpenAI    & GPT3.5& Embed  & 60.32\% & 49.55\% \\
\midrule
OpenAI    & GPT3.5& GEM-RAG (k-Means)   & 61.42\% & 50.83\%\\
\bottomrule
\end{tabular}
\caption{Results on the QuALITY dev dataset for all embedding, LLM and RAG pairs.}
\label{tab:exp_qual}
\end{table*}

\begin{remark}
Suppose the document includes $k$ essential themes. The modularity of $S$ can be formulated as follows:
$$
S = \left[
\begin{array}{ccc}
\hat{S}_{11} & \cdots & \hat{S}_{1k}  \\
\vdots & \ddots & \vdots \\
\hat{S}_{k1} & \cdots & \hat{S}_{kk} \\
\end{array}
\right]
$$
where $\hat{S}_ij$ has $n_i$ rows. Thus, each diagonal block $\hat{S}_{ii}$ satisfies $0 < n_i - ||\hat{S}_{ii}||_F < \epsilon$, which implies $||\hat{S}_{ij}||_F \rightarrow 0$ for each off-diagonal block. Then we can characterize $S$'s spectrum via the behavior of its eigenvalues, i.e., $ 0 < 1 - \lambda_i < \epsilon, \forall i \in \{1, 2, \ldots, k\}$ and $0 < |\lambda_i| < \epsilon, \forall i \in \{k+1, k+2, \ldots, n\}$.
\end{remark}

These properties confer on $S$ effective clustering capabilities, where text chunks within the same theme exhibit higher similarity and lower similarity across different themes.

\begin{remark}
    Consider a sequence where each term is the ratio $\beta_i$ of $\lambda_i, \forall i \geq 2$ to the largest eigenvalue $\lambda_1$. If there exists some index $d \geq 2$ such that $\beta_d$ is close to 1, and $\beta_d - \beta_{d+1} > c$ where $c$ indicates the cutoff that identifies the first significant gap between a pair of adjacent ratios, then $d$ is the estimated number of essential themes in the document.
\end{remark}

\begin{remark}
    Let $\Lambda = \sum_i \lambda_i^2$. A higher $\Lambda$ indicates more distinct clustering themes within the document; otherwise, the differentiation between themes becomes ambiguous.
\end{remark}

\subsubsection{Retrieval}
Given a built GEM, our method is then ready to answer prompts/question about the given dataset. The process works as follows, given a prompt/question $p$, and some budget $\mathcal{B}$ of nodes to return, we first produce an embedding of the prompt to give $v_p = E(p)$. We then find, out of the entire graph, the utility question that has the highest $\mathtt{SIM}$. Specifically, let $\mathcal{Q} = \{q_1, q_2, \ldots, q_n\}$ be the set of all utility questions, find $q^* = \underset{q \in \mathcal{Q}}{\mathrm{argmax}} \, \mathtt{SIM}(v_p, E(q))$. 
Then, from the associated node with $q^*$ we perform a best first search, to find the next nodes, up to $\mathcal{B}$ to include in the context set. Once we reach the budget we return the context. The full details can be seen in Algorithm \ref{alg:retrieval}.

The LLM is then given this context, followed by the question and prompted to answer. 

\subsection{Method Trade-Offs}
While the robustness of our method leads to our improved results, it does come with some trade-offs that are important to discuss, particularly in terms of computational complexity, and potential costs. 

First, generating utility questions via an LLM can become a significant cost, either computationally or monetarily ,depending on the number of nodes in the dataset, and the number of utility questions. Generating a graph with $n$ nodes and $q$ utility questions requires $nq$ LLM calls. The graph building is all pre-computed, so in most cases this extra cost is okay, but it is worth noting. 

Secondly, in order to generate higher level summary nodes we do eigendecomposition which  has a complexity of $O(n^3)$ where $n$ is the number of nodes. With a large number of nodes this complexity may require consideration.


\section{Experiments}
Our method, and RAG methods in general can be used for a host of tasks, including AI agent memory, fine-tuning LLMs for niche scientific domains etc. For purposes of comparison we evaluate the efficacy of our model in the context of QA for medium length documents. 

\subsection{Setting}
It should be noted that RAG methods in general aim to study the specific problem of retrieving optimal small context windows from large data sets and returning them to an LLM for processing. However, this setting is of ongoing debate and change within the research community, as large LLMs have increasingly large context lengths, making it possible to fit massive amounts of data within a single prompt. We feel, that even with larger and larger context windows, the problem of retrieval is still of interest because in a real-world setting, the data sets of interest are often still much larger than the current context windows of the biggest models, and returning massive amounts of context tokens is impractical and costly. Secondly, even with a larger context, it has been shown that LLMs can be prone to forget information in the context, and can be prone to hallucinate information. Lastly, with more precise retrieval it makes it easier to verify and track the information that an LLM is using to answer a question. For these reasons, we feel that the retrieval methods are still highly motivated despite the evergrowing context length of large models. Retrieval also requires no finetuning, and uses LLMs ``off-of-the-shelf". 

To properly compare our method to other recent work we study datasets where often the entire data set \textit{can} fit in the context of an LLM. For this reason, while our method outperforms other retrieval methods, which are all limited to 400 tokens of context, there has been work that uses thousands of tokens of context and outperforms retrieval metrics such as \cite{zeroscrolls}. Again, we feel this is impractical for larger data sets, and only use these datasets to showcase our method against similar recent work.

\subsection{Datasets}
\paragraph{QuALITY}
The Question Answering with Long Input Texts, Yes! (QuALITY) dataset \cite{pang2022quality} contains 230 medium length passages (about 5000 tokens), for which each passage has associated multiple choice questions and ground truth answers for. Additionally, each set of questions has a subset of \textit{HARD} questions which are particularly challenging. Specifically we use the `dev' data split. For the ablations experiments we use the first 50 passages from this set, whereas for the main experiments in Table \ref{tab:exp_qual} we use the remaining 180. 

\paragraph{Qasper}
The Qasper data set \cite{qasper_dasigi-etal-2021-dataset} contains over 1500 academic NLP papers. Each paper has associated multiple choice questions and ground truth answers. We perform our experiments over the first 100 papers in the data set. 

\subsection{Baselines}
We compare our method to the standard RAG method, of embedding each chunk of text, and embedding the given prompt, and finding the most similar chunks to the prompt, up to a specific budget. Additionally, we compare to a recent work that showed promising results, Recursive Abstractive Processing
For Tree-Organized Retrieval (RAPTOR) method \cite{raptor}. RAPTOR primarily aims to tackle the problem of producing hierarchies of nodes, that summarizes the text passage  appropriately. 

\begin{figure*}[t]
    \centering
    \includegraphics[width=.99\textwidth]{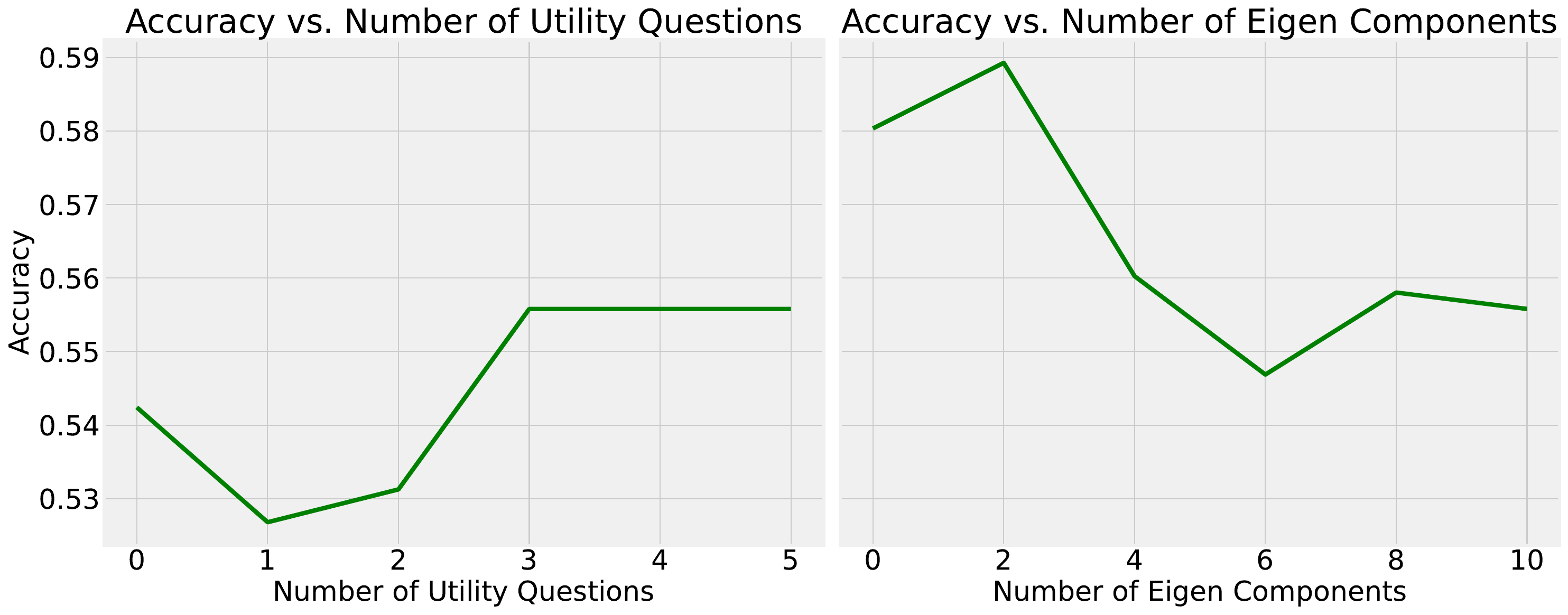}
    \caption{Ablation study on the effect of number of utility questions and eigencomponents for a 50 passage subset of the QuALITY data set. The accuracy from more utility questions quickly becomes saturated, whereas the best number of eigencomponents is two. Intuitively the number of utility questions capture the amount of information per chunk, whereas the number of components covers the importance of summaries in the domain of questions.}
    \label{fig:abl}
\end{figure*}

\subsection{Evaluation}
The QuALITY dataset has multiple choice questions with ground truth answers, for this dataset we calculate the accuracy and report the overall accuracy, as well as the accuracy on the HARD subset. The Qasper dataset has answer types that are either Abstractive, Extractive, Yes/NO or Answerable/Unanswerable. We use the F1 score to evaluate efficacy of the methods for this dataset.

\begin{table}[t]
\centering
\begin{tabular}{lccc}
\toprule
Embedding & LLM   & RAG    & F1   \\
\midrule
SBERT     & GPT3.5& GEM-RAG   & 18.53\%  \\
SBERT     & GPT3.5& RAPTOR & 18.51\%  \\
SBERT     & GPT3.5& Embed  & \textbf{19.24}\% \\
\midrule
OpenAI    & GPT3.5& GEM-RAG   & \textbf{20.13}\%\\
OpenAI    & GPT3.5& RAPTOR & 18.13\%  \\
OpenAI    & GPT3.5& Embed  & 19.07\%  \\
\bottomrule
\end{tabular}
\caption{Results on the Qasper data set for all embedding, LLM and RAG pairs.}
\label{tab:exp_qasper}
\end{table}

\subsection{Experimental Parameters}
Our objective with these experiments is to evaluate the efficacy of our model across different text embedding models and LLMs. We use the SBERT \cite{sbert}, and the OpenAI text-embedding-ada-002 text embedding models. We also use the UnifiedQA \cite{unifiedqa-test} and GPT3.5 Turbo LLMs. For the QuALITY data set we consider all possible combinations of text encoders, LLMs and RAG methods. For the Qasper dataset we consider all embedding and RAG models but only using GPT-3.5 Turbo as the LLM. 

For all experiments we use a chunk size of 100 tokens, and we allow 400 tokens of context, meaning four nodes of context. Even though GPT3.5 can support a much larger context, we aim to study the setting where only few nodes/chunks can be used, to better isolate the effectiveness of the RAG method in question, rather than the attention mechanism of the LLM. For the summarization method for GEM-RAG and RAPTOR, as well at the utility question method for GEM-RAG we use GPT3.5 Turbo as the LLM.
All similarity measurements were done using cosine similarity. For the GEM-RAG method we use two eigencomponents, and five utility questions. In our ablation experiments we show the effect of varying both of these parameters. For the ablation experiments we evaluate the accuracy on the first 50 articles of the QuALITY data set, whereas for the main experiments we evaluate on the latter 180 articles.

\subsection{Results}
\textbf{QuALITY Results}
The results from our experimentation can be seen in Table \ref{tab:exp_qual}. We observe that in all settings, except for those using the OpenAI embedding model, and UnifiedQA LLM, our model gives the best performance, in both the over all accuracy and accuracy on the HARD subset. Also notably, the difference is exaggerated most on the highest accuracy setting, using OpenAI's embedding, and GPT-3.5 Turbo, suggesting this difference is exaggerated the more robust the embedding and LLM models used are.

Additionally, we looked at the performance of our method if we use k-Means instead of spectral clustering. We observe a performance drop, which indicates the spectral graph analysis performs better than more standard clustering. 

\textbf{Qasper Results}
The results from our experimentation can be seen in Table \ref{tab:exp_qasper}. In this setting we only use GPT3.5Turbo as the LLM and consider using both SBERT and the OpenAPI text embedding models. We see that our method performs best for the OpenAI's text-embedding-ada-002 embedding model, and gives the best over all score. However, it performs behind the standard RAG method when considering SBERT. 

\textbf{GEM For Data Visualization} We would like to stress that while GEM has been formulated to be primarily used as a method for RAG, the produced GEM is a standalone object that can be used with any LLM to do QA work, as well as a tool to visualize and organize data. We provide an example visualization for a single story in the QuALITY data set at the following url: \url{https://detailed-swan.static.domains/GEM.html}.

\textbf{Ablation Study}
In order to test the tuneable hyperparameters of our model, the number of eigencomponents and the number of utility questions, we performed three ablation experiments, the results of which can be seen in Figure \ref{fig:abl}. In the first we keep the number of components constant at ten, and vary the number of utility questions. We observe that the accuracy with respect to the number of utility questions increased up until three, but quickly becomes saturated. Zero utility questions indicates that we just compare the embedding of each text chunk directly. Conversely, for the second plot, we hold the number of utility questions at five, and vary the number of eigencompnents. Here we observe the actual best performance is with two components. 

Intuitively, the number of eigencomponents capture how important high-level themes are to the questions of interest, and the number of utility questions capture how much detail can be in embedded in a single chunk. These parameters should be tuned to the specific task at hand, and we used 50 passages from QuALITY as a hold out set to inform the hyperparameters for the remaining 180 passages.

In the third study, seen in Table \ref{tab:percent_eigen} we look at, for the number of eigencomponents available, the percent of nodes returned as context that are eigen summary nodes. As can be seen in the table, with more eigen-components, they become more likely to be returned as part of the context. This may be good for questions that are more high-level, but bad for more detailed questions, as is the case in the QuALITY dataset. 


\begin{table}[t]
\centering
\begin{tabular}{lc}
\toprule
Eigen-components & Percent of nodes returned that are eigen/summary \\
\midrule
0 & 0.0 \\
2 & 14.4 \\ 
4 & 24.5 \\ 
6 & 32.8 \\ 
8 & 38.1 \\ 
10 & 42.6 \\ 

\bottomrule
\end{tabular}
\caption{Percent of returned nodes that are eigen/summary nodes.}
\label{tab:percent_eigen}
\end{table}

\section{Discussion and Conclusion}
We developed GEM-RAG, a method for RAG inspired by human cognition, that tags each memory or chunk by the specific utility of its information and relation to other memories. Further, we use these utility questions to formulate a weighted fully connected graph. We perform an eigendecomposition on this memory graph to robustly extract ``eigenthemes'', and create summary nodes for each theme. We observe in most cases, for multiple text embeddings and LLMs, our method out-performs standard baselines. We also show that a produced GEM is a standalone object: it can be used with any LLM to be searchable and conversable, and provides a principled visualization for understanding textual data. We believe an optimal RAG method has the ability to greatly improve the ability of LLMs, enabling real AI agents that can leverage massive histories of conversations, or adapt to massive niche data sets without fine-tuning. Further, these RAG methods can extend to LMMs, retrieving text, images, videos, sound, etc., and bringing us closer to simulating human cognition. 

\section{Acknowledgements}
This project is partially supported by the National Science Foundation (NSF); the Eric and Wendy Schmidt AI in Science Postdoctoral Fellowship, a program of Schmidt Sciences, LLC; the National Institute of Food and Agriculture (US-DA/NIFA);  the Air Force Office of Scientific Research) (AFOSR), and Toyota Research Institute (TRI).
\bibliographystyle{IEEEtran}
\bibliography{main}

\end{document}